\documentclass[conference]{IEEEtran}
\IEEEoverridecommandlockouts

\usepackage{cite}
\usepackage{amsmath,amssymb,amsfonts}
\usepackage{algorithm}
\usepackage[noend]{algpseudocode}
\usepackage{graphicx}
\usepackage{textcomp}
\usepackage{array}
\usepackage{csquotes}
\usepackage{multirow}
\usepackage{xcolor}
\usepackage{makecell}
\usepackage{subcaption}
\usepackage[hidelinks]{hyperref}
\usepackage{stfloats}
\usepackage{fancyhdr}
\usepackage{lettrine}
\definecolor{my_blue}{RGB}{0,0,255}

\newcolumntype{C}[1]{>{\centering\arraybackslash}p{#1}}

\begin{document}

\title{Optimally Deep Networks -- Adapting Model Depth to Datasets for Superior Efficiency}

\author{
\hspace{1.6cm}\IEEEauthorblockN{Shaharyar Ahmed Khan Tareen}
\hspace{1.6cm}\textit{University of Houston}\\
\hspace{1.6cm}Houston, TX, USA \\
\hspace{1.6cm}stareen@cougarnet.uh.edu
\and
\hspace{2.0cm} \IEEEauthorblockN{Filza Khan Tareen}
\hspace{2.0cm} \textit{National University of Sciences and Technology}\\
\hspace{2.0cm} Islamabad, Pakistan \\
\hspace{2.0cm} ftareen.bse2021mcs@student.nust.edu.pk
}
\maketitle

\thispagestyle{fancy}
\fancyhf{}
\fancyfoot[C]{\textbf{Preprint:} Accepted at the 2025 5\textsuperscript{th} International Conference on Digital Futures and Transformative Technologies (ICoDT2)}\renewcommand{\headrulewidth}{0pt}
\renewcommand{\footrulewidth}{0pt}

\begin{abstract}
Deep neural networks (DNNs) have provided brilliant performance across various tasks. However, this success often comes at the cost of unnecessarily large model sizes, high computational demands, and substantial memory footprints.
Typically, powerful architectures are trained at full depths but not all datasets or tasks require such high model capacity. Training big and deep architectures on relatively low-complexity datasets frequently leads to wasted computation, unnecessary energy consumption, and excessive memory usage, which in turn makes deployment of models on resource-constrained devices impractical. To address this problem, we introduce the concept of Optimally Deep Networks (ODNs), which provides a balance between model depth and task complexity. Specifically, we propose a NAS like training strategy called \enquote{progressive depth expansion}, which begins by training neural networks at shallower depths and incrementally increases their depth as the earlier blocks converge, continuing this process until the target accuracy is reached. ODNs use only the \enquote{optimal} depth for the tasks at hand, removing redundant layers. This cuts down future training and inference costs, lowers the model’s memory footprint, enhances computational efficiency, and facilitates deployment on edge devices. Empirical results show that the optimal depths of ResNet-18 and ResNet-34 for MNIST and SVHN, achieve up to 98.64 \% and 96.44 \%  reduction in memory footprint, while maintaining a competitive accuracy of 99.31 \% and 96.08 \%, respectively.
\end{abstract}
\vspace{0.15cm}

\section{Introduction}

\noindent Deep Neural Networks (DNNs) have become the cornerstone of modern machine learning, due to their brilliant performance across various domains \cite{DL_1,DL_2,DL_3}. However, these feats come at the expense of large model sizes due to deeper architectures and correspondingly higher computational and memory demands. Although large models have proven essential for complex datasets (ImageNet-1K, ImageNet-21K, large-scale NLP datasets etc.) \cite{DL_2}, not all tasks have such high level of complexity. Many domains involve tasks that are significantly simple or have lower complexity such as some datasets in medical imaging \cite{MI_1}, industrial inspection \cite{II_1}, digit or character classification \cite{CC_1,DC_1}, and image matching \cite{SN_1}. Using deep architectures (ResNet-18, ResNet-34, ResNet-50, ResNet-101 etc.) in simple scenarios leads to inefficiencies, including wasted computation/time, increased latency during inference, unnecessary energy consumption, and excessive memory footprint, hindering their deployability on resource-constrained devices e.g. mobile phones and edge devices.

\vspace{0.20cm}
Many strategies for efficient deep learning \cite{EDL_1,EDL_2} have been proposed including pruning \cite{pruning}, quantization \cite{quantization}, knowledge distillation \cite{KD_1,KD_2,KD_3,KD_4,KD_5}, dynamic inference \cite{DI_1}, and Neural Architecture Search (NAS) \cite{NAS_1,NAS_2,NAS_3}. Pruning and quantization compress the networks but cannot substantially reduce their size on disk. Knowledge distillation transfers knowledge from a large teacher model to a smaller student model, however, it does not provide the optimized architecture for the task at hand. Dynamic inference techniques adaptively skip layers in a model during inference to reduce the latency but fail to address the problem of large memory footprint. Neural Architecture Search (NAS) provides an architecture search process, however, it has huge computation cost leading to large carbon footprint \cite{OFA_1}, extreme time consumption (as it typically trains thousands of candidates to find the optimum architecture), and the complexity of defining the search space, which may sometimes miss powerful architectures.

\begin{figure}[!t]
    \centering
    \vspace{0.08cm}
    \includegraphics[width=1.0\columnwidth]{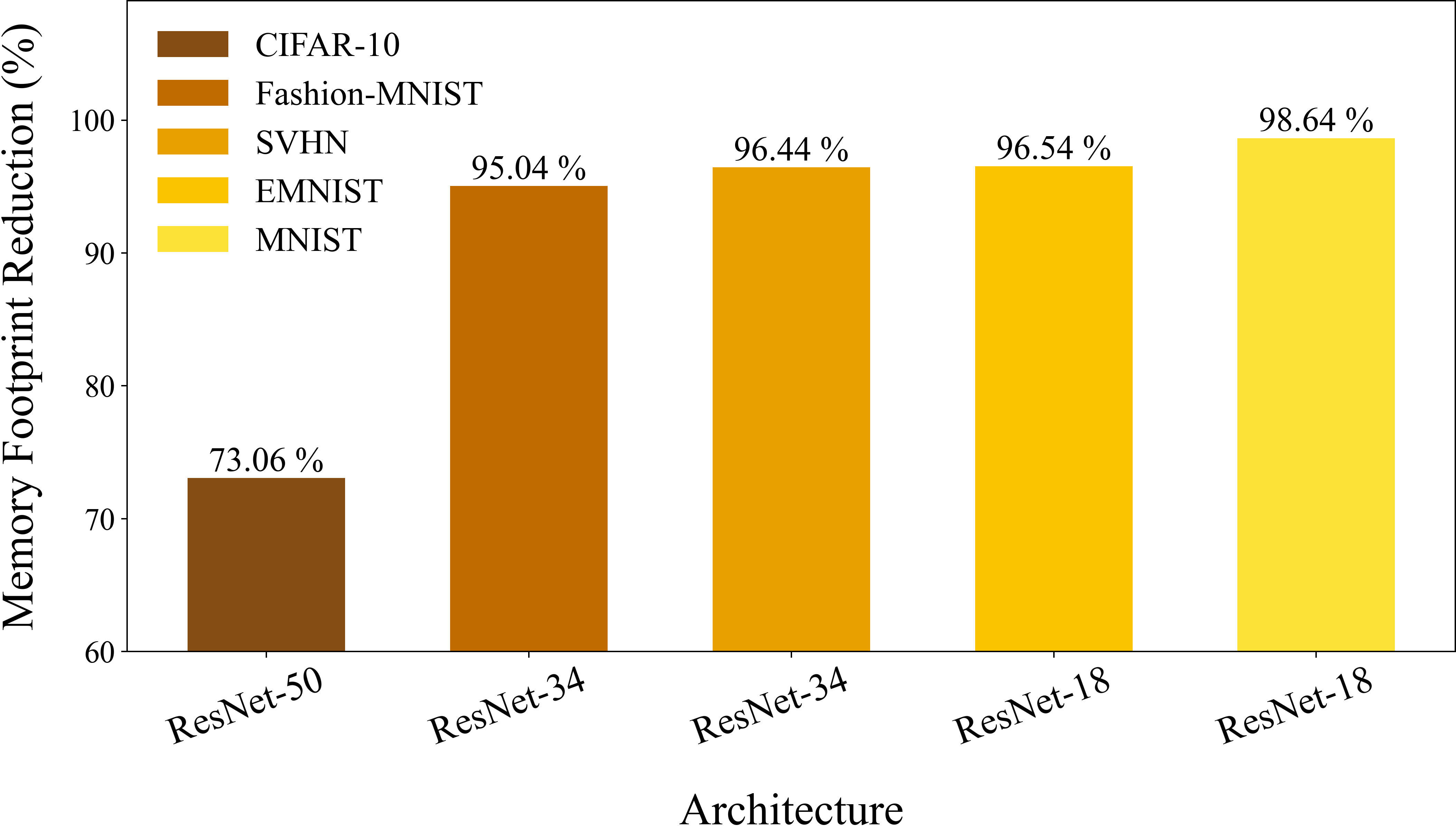}
    \caption{\centering Memory footprint reduction of networks at their optimal depths across five datasets (The accuracies of the ODNs remain within 1 \% of the full-depth networks).}
    \vspace{-0.20cm}
    \label{fig:figure_1}
\end{figure}

\vspace{0.20cm}
We introduce the concept of Optimally Deep Networks (ODNs), which use a simple and smart NAS like training strategy that enables adapting the depth of a model to the complexity of the task at hand. Instead of directly training the model with full depth, \enquote{progressive depth expansion} is applied that begins by activating only the shallow layers and incrementally deepens the network once the earlier layers have converged, continuing this process until the desired performance is achieved. When the target performance is achieved with a depth shallower than the original architecture, the Optimally Deep Network is extracted and fine-tuned. This depth-search strategy enables networks to allocate only the necessary depth capacity for the tasks, reducing substantial memory footprint, future training costs, computation overheads, and inference run-time, all without a significant compromise on accuracy. Once the optimal depth of an architecture is obtained, it can be used to efficiently deploy the model at a large scale, reducing resource wastage. The same depth can be reused for future trainings on datasets of similar complexity, avoiding the cost of performing another search. Moreover, the ODNs can be further compressed using pruning and quantization techniques.

\vspace{0.20cm}
Unlike Neural Architecture Search (NAS), progressive depth expansion is a simple depth search strategy which does not require defining a complex search space. It can be straightforwardly applied on many powerful architectures such as Residual Networks (ResNet-18, ResNet-24, ResNet-50, ResNet-101 etc.) \cite{DL_2}, Wide Residual Networks (WRN-40-2, WRN-16-8, WRN-28-10 etc.) \cite{WRN}, Mobile Networks (MobileNetV1, MobileNetV2, MobileNetV3 etc.) \cite{MN_1,MN_2} and many more, by dividing them into different depth levels (blocks) and progressively training them from shallower to deeper blocks until the desired performance is achieved with the optimal depth. ODNs offer a new perspective on efficient deep learning by searching optimal depth of well-known networks according to the complexity of tasks, bridging the gap between performance and efficiency.

\vspace{0.25cm}
\noindent \textbf{Key Contributions:} Our contributions are highlighted below:
\begin{itemize}
    \item We introduce Optimally Deep Networks (ODNs), in which \enquote{model depth} is adapted according to the complexity of task at hand, with competitive performance.
    \item We propose a novel training paradigm called \enquote{progressive depth expansion} that gradually deepens the model-depth during training to achieve the target accuracy.
	\item We show that ODNs significantly reduce memory footprint and inference costs, enabling efficient deployment on edge and resource-limited devices.
	\item We empirically demonstrate the effectiveness of ODNs on five datasets: MNIST, EMNIST, SVHN, Fashion-MNIST, and CIFAR-10, by optimizing the depths of three architectures: ResNet-18, ResNet-34, and ResNet-50.   
	\item ResNet-18 trained through progressive depth expansion on MNIST provides \textbf{98.64 \% reduction} in model size (from 44.78 MB to 0.61 MB) while preserving competitive performance with 99.39 \% accuracy. Similarly, ResNet-34 trained on SVHN provides \textbf{96.44 \% reduction} in model size (from 85.29 MB to 3.04 MB) with a competitive accuracy of 96.08 \%.
	\item The code of ODNs is released at the following link: \href{https://github.com/saktx/Optimally_Deep_Networks}{\textbf{https://github.com/saktx/Optimally\_Deep\_Networks}}.
\end{itemize}
 
\vspace{0.15cm}
\section{Related Work}
The pursuit of greater efficiency in deep learning inspired researchers to introduce various approaches aimed at cutting down the computational costs, memory footprint, inference latency and other deployment barriers of the models, while retaining their performance \cite{EDL_1,EDL_2}. Most popular research directions related to efficient deep learning include Neural Architecture Search (NAS), model compression techniques (sparsification, pruning, and quantization), knowledge distillation, dynamic inference, and Once-For-All (OFA) training.

\begin{figure}[!t]
	\vspace{-0.05cm}
    \centering
    \includegraphics[width=0.35\textwidth]{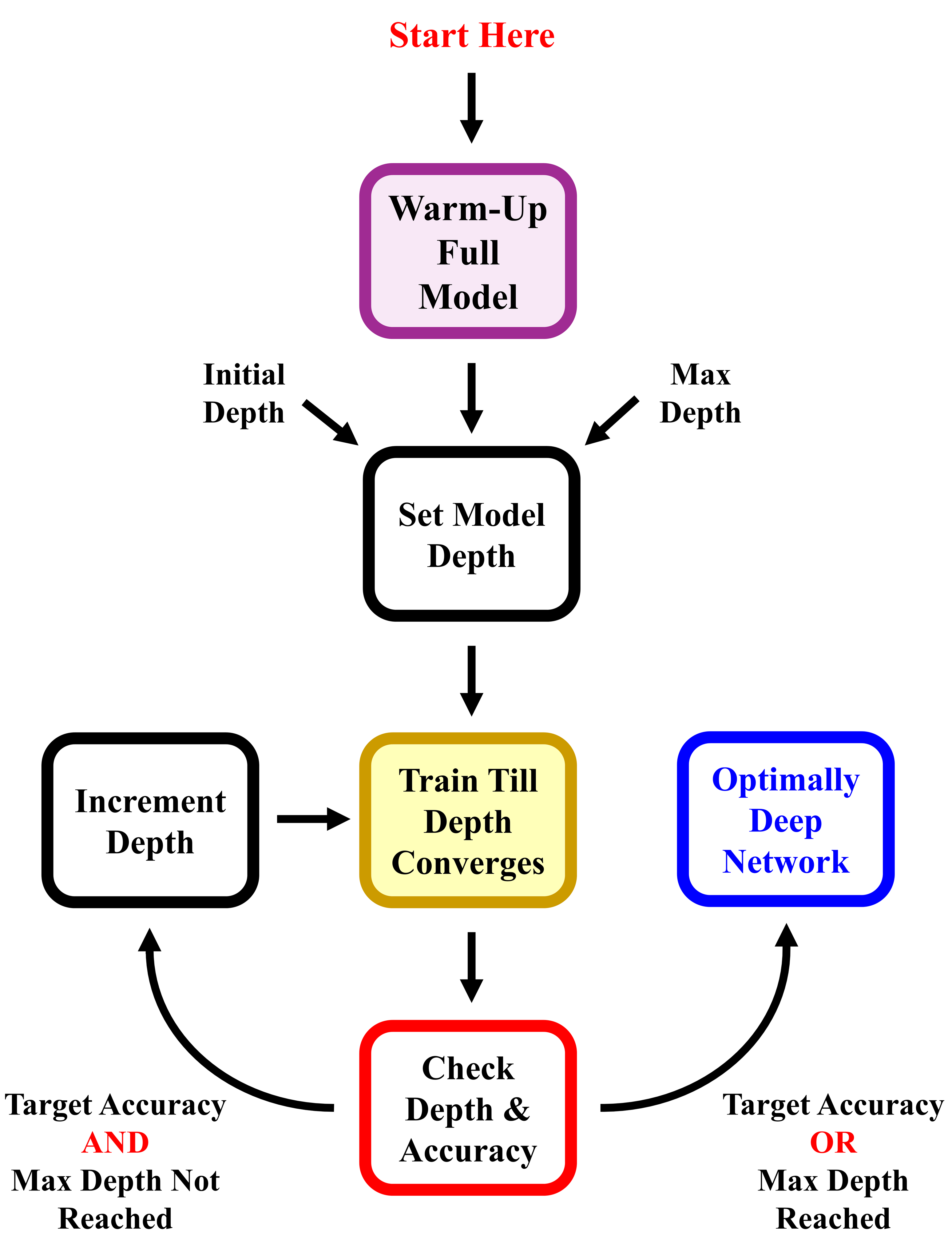}
    \caption{\centering Procedure for obtaining Optimally Deep Networks.}
    \vspace{-0.2cm}
    \label{fig:figure_2}
\end{figure}

\subsection{Neural Architecture Search (NAS)}
NAS automated the process of finding a neural network that is optimized with respect to accuracy, robustness, and/or efficiency by exploring a vast search space of possible architectures without requiring extensive evaluation of trial-and-error manually \cite{NAS_1,NAS_2,NAS_3}. NAS has been able to provide high performing architectures by leveraging reinforcement learning such as RENAS \cite{RENAS}, gradient based optimization like DARTS \cite{DARTS}, and evolutionary search algorithms e.g. NPENAS \cite{NPENAS}. However, the search process of NAS can be computationally intensive, requiring up to thousands of GPU hours and since NAS generated architectures are not universally optimal across all the datasets, a new architecture search is required for each task, undermining the scalability \cite{NAS_1}. This limitation makes NAS extremely costly for such scenarios.

\begin{figure*}[!t]
    \centering
    \includegraphics[width=1.0\textwidth]{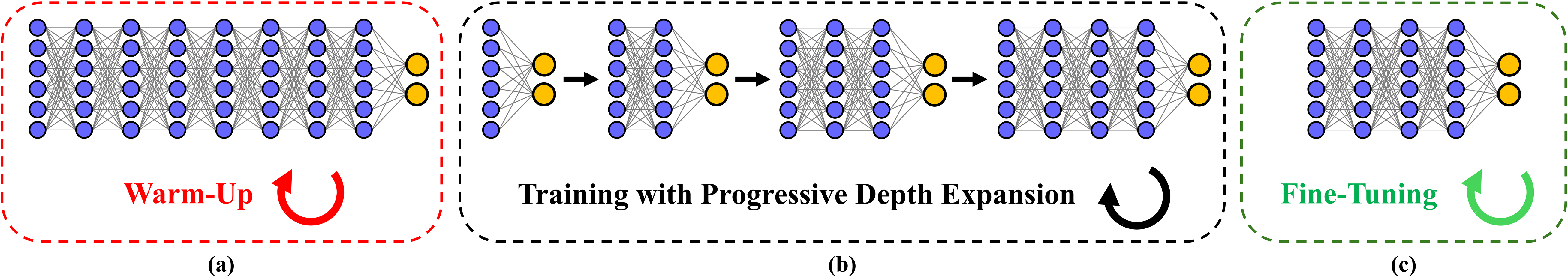}
    \caption{\centering Training and depth search process of Optimally Deep Networks (ODNs): (a) The selected network is first warmed up by training for a few epochs with a small learning rate for all the depth levels. (b) Progressive depth expansion is applied during training, starting with a shallow depth and incrementally adding deeper blocks once the shallower ones have fully converged. (c) The network searched with the optimal depth is then fine-tuned to maximize its performance.}
    \label{fig:figure_3}
\end{figure*}

\subsection{Model Compression Techniques}
Sparsification, pruning, and quantization are widely used techniques for compressing large neural networks \cite{pruning,quantization}. Sparsification encourages the model to use as few parameters as possible by pushing their magnitudes toward zero during training. Subsequently, pruning is applied to remove these zeroed-out weights or neurons. Quantization on the other hand reduces the numerical precision of model parameters to decrease its memory footprint \cite{quantization}. Although these compression methods reduce the number of parameters or their numerical precision, they do not provide substantial reduction in memory footprint, particularly in the case of unstructured sparsification. Moreover, pruning and quantization are typically applied post training, meaning the model is first trained at full depth before compression, still leading to larger memory footprints because the model's structure remains unchanged.

\subsection{Once-for-All Training Frameworks}
Once-for-All (OFA) training frameworks aim to train a large overparametrized super-net only once, that can provide multiple optimized sub-networks of varying sizes later, facilitating flexible deployment across diverse hardware platforms \cite{OFA_1,OFA_2,OFA_3}. Following this approach, SNNs \cite{SNN_1,SNN_2} train a super-net for different widths, enabling accuracy and efficiency trade-off. Although OFA training provides flexibility and avoids retraining for each device, it introduces significant training complexity and does not perform well across different datasets. Moreover, the super-net parameters are shared among all the sub-nets that leads to challenges in optimization, high training costs, and reduced specialization of the sub-nets compared to individually trained models \cite{SNN_1,OFA_1}.

\vspace{0.20cm}
While all these methods have advanced the domain of efficient deep learning, each one suffers from its limitations which highlight the need for an alternative and simple paradigm that directly adapts model size to the complexity of the task at hand. Optimally Deep Networks (ODNs) tackle this gap by progressively expanding model depth in a data-driven manner, cutting down the unnecessary memory footprint while maintaining competitive accuracy.

\vspace{0.15cm}
\section{Optimally Deep Networks (ODNs)}
We propose \textbf{Optimally Deep Networks (ODNs)}, which adapt model-depth to the complexity of the dataset via \enquote{progressive depth expansion}. Instead of training the model with full-depth, we first divide its depth in partitions and assign an output layer for each depth. Next, we warm up the model by training it for few epochs at all the depth levels, with a small learning rate. This stabilizes the optimization process for all depth partitions of the model. The warmed-up model parameters are saved for assisting in the search of optimal depth later. Afterwards, the user specifies a \enquote{target accuracy} according to the desired performance requirements. The progressive depth expansion based training starts by first activating a shallow portion of the network that only contains its initial block(s). Once this shallow portion is converged, additional blocks are appended, and training is  restarted using the warmed-up parameters (till this new depth of the network) that were saved after the warm-up. This process continues until the model achieves the target accuracy or maximum depth is reached. If the target accuracy is obtained with a depth smaller than the model’s original depth, the current depth is returned as the \textbf{\enquote{optimal depth}} for which the ODN is extracted and fine-tuned. This dynamic depth search approach ensures that depth capacity of the network is allocated for the dataset as needed, preventing over-parameterization and reducing the model's memory footprint without compromising accuracy. 

\vspace{0.20cm}
Algorithm~\ref{alg:algo_1} outlines the procedure of training Optimally Deep Networks and the workflow to obtain ODNs is shown in Figure~\ref{fig:figure_2}. In the following subsections, we describe the key phases involved in obtaining an Optimally Deep Network.

\begin{algorithm}[!b]
\caption{\centering \textbf{:} Training Optimally Deep Networks (ODNs)}
\label{alg:algo_1}
\begin{algorithmic}[1]
\Require training dataset $\mathcal{S}$, neural network $\mathcal{N}$, depth levels $\mathcal{D}=\{1, 2, 3, \dots, \mathcal{D}_{\max}\}$, warm-up epochs $E_{w}$, stopping epochs $E_{stop}$, warm-up learning rate $\alpha_{w}$, learning rate $\alpha$, initial depth $\mathcal{D}_{i}$, final depth $\mathcal{D}_{f}$, target accuracy $\mathcal{A}_{tar}$
\Ensure trained network $\mathcal{N}_{\Theta}$ with optimal depth $\mathcal{D}_{opt}$
\State Activate $\mathcal{N}$ with depth $\mathcal{D}_{\max}$
\State Initialize network parameters $\Theta$
\State Initialize $optimizer$ with $\Theta$ and $\alpha_{w}$
\For{$e = 1$ to $E_{w}$}
    \State Clear gradients: $optimizer.zero\_grad()$
    \State Sample batch $(x, y)$ from $\mathcal{S}$
    \State Compute loss: $\mathcal{L}(y, \mathcal{N}(x))$
    \State Backpropagate: $\mathcal{L}.backward()$
    \State Update parameters: $optimizer.step()$
\EndFor
\State \textbf{Save:} Save parameters of $\mathcal{N}$ as $\Theta_{w}$
\State Activate $\mathcal{N}$ with depth $\mathcal{D}_{i}$
\State Set learning rate $\alpha$
\State Initialize $A_{best} \gets 0$, $A_{current} \gets 0$, $E_{current} \gets 0$, $E_{stop} \gets 23$, $\mathcal{D}_{current} \gets \mathcal{D}_{i}$
\While{$A_{current} \le A_{target} \ \textbf{and} \ \mathcal{D}_{current} \le \mathcal{D}_{f}$}
\State $E_{current} \mathrel{+}= 1$
	\For{\textbf{each} batch $(x, y)$ \textbf{in} $\mathcal{S}$}
    \State Clear gradients: $optimizer.zero\_grad()$
    \State Sample batch $(x, y)$ from $\mathcal{S}$
    \State Compute loss: $\mathcal{L}(y, \mathcal{N}(x))$
    \State Backpropagate: $\mathcal{L}.backward()$
    \State Update parameters: $optimizer.step()$
    \State Evaluate $\mathcal{N}$ till $\mathcal{D}_{current}$ to get $A_{current}$
    \EndFor
    \If{$A_{current} > A_{best}$}
    \State \textbf{Save:} Save best model $\mathcal{N}_{\Theta}$
    \State $A_{best} = A_{current}$
    \State $E_{current} = 0$
	\EndIf
	
    \If{$E_{current} \bmod 5 = 0$}
    \State $\alpha \gets 0.6 \times \alpha$
	\EndIf
	
	\If{$E_{current} = E_{stop}$}
    \State $D_{current} \mathrel{+}= 1$
    \State \textbf{Load:} Load parameters of $\mathcal{N}$ from $\Theta_{w}$
	\EndIf
\EndWhile
\State $\mathcal{D}_{opt} \gets \mathcal{D}_{current}$
\State \Return $\mathcal{N}_{\Theta}$, $D_{opt}$ 
\end{algorithmic}
\end{algorithm}

\begin{table*}[t]
\definecolor{my_blue}{RGB}{0,0,255}
\renewcommand{\arraystretch}{1.3}
\centering
\footnotesize
\caption{\centering Comparison of Model Depth, Parameter Count, Memory Footprint (Float32 Precision), FLOPs, and Performance of Standard Networks with Full Depths versus Corresponding Optimally Deep Networks (ODNs). Reduced Parameter Counts Show Significant Reduction in Memory Footprints whereas Reduced FLOPs Indicate Improvements in Inference Speed.}
\begin{tabular}{|C{1.8cm}|C{2.4cm}|C{1.2cm}|C{1.9cm}|C{1.9cm}|C{2.5cm}|C{1.45cm}|C{1.45cm}|}
\hline
\textbf{DATASET} & \textbf{Architecture} & \textbf{Depth} & \textbf{Parameters $\downarrow$} & \textbf{Size on Disk $\downarrow$} & \textbf{\% Size Reduction $\uparrow$} & \textbf{FLOPs $\downarrow$} & \textbf{Accuracy $\uparrow$} \\
\hline

\multirow{2}{*}{MNIST} & ResNet-18 \phantom{$_{8/8}$} & 8/8 & 11.17 M & 44.78 MB & \multirow{2}{*}{98.64 \%} & 458.21 M & 99.61 \% \\
& ResNet-18 $_{2/8}$ & 2/8 & 0.15 M & \textcolor{my_blue}{\textbf{0.61 MB}} & & 117.36 M & 99.31 \% \\		
\hline

\multirow{2}{*}{EMNIST} & ResNet-18 \phantom{$_{8/8}$} & 8/8 & 11.17 M & 44.78 MB &\multirow{2}{*}{96.54 \%} & 458.21 M & 95.17 \% \\
& ResNet-18 $_{3/8}$ & 3/8 & 0.38 M & \textcolor{my_blue}{\textbf{1.55 MB}} & & 162.37 M & 95.04 \% \\		
\hline

\multirow{2}{*}{SVHN} & \phantom{*}ResNet-34 \phantom{$_{6/16}$} & 16/16 & 21.28 M & 85.29 MB &\multirow{2}{*}{96.44 \%} & 939.91 M & 96.51 \% \\
& \phantom{*}ResNet-34 $_{5/16}$ & 5/16 & 0.75 M & \textcolor{my_blue}{\textbf{3.04 MB}} & & 365.63 M & 96.08 \% \\	
\hline

\multirow{2}{*}{Fashion-MNIST} & \phantom{*}ResNet-34 \phantom{$_{6/16}$} & 16/16 & 21.28 M & 85.29 MB &\multirow{2}{*}{95.04 \%} & 939.91 M & 94.52 \% \\
& \phantom{*}ResNet-34 $_{6/16}$ & 6/16 & 1.04 M & \textcolor{my_blue}{\textbf{4.23 MB}} & & 337.08 M & 93.54 \% \\	
\hline

\multirow{2}{*}{CIFAR-10} & \phantom{**}ResNet-50 \phantom{$_{16/16}$} & 16/16 & 23.52 M & 94.43 MB &\multirow{2}{*}{73.06 \%} & 1315.02 M & 95.01 \% \\
& \phantom{**}ResNet-50 $_{11/16}$ & 11/16 & 6.31 M & \textcolor{my_blue}{\textbf{25.44 MB}} & & 906.19 M & 94.12 \% \\	
\hline
\end{tabular}
\label{tab1}
\end{table*}

\subsection{Depth Partitioning \& Multiple Output Layers Setup}
To train a model at different depths, its depth is partitioned into different levels. When a depth is activated, the data  passes only through the activated depth to provide the output. We divide the depth of ResNet-18, ResNet-34,  and ResNet-50 into 8, 16, and 16 blocks, respectively. The partitions are based on number of residual blocks. An activated depth level of 4 means that only the first 4 blocks of the model are active. Each block in ResNet-18 and ResNet-34 contains two convolutional layers each followed by a BatchNorm layer, whereas, ResNet-50 blocks contain three convolutional layers (each followed by a BatchNorm layer). A separate output layer is employed in the network architecture for each depth level (to facilitate data flow), meaning for 16 depth levels there will be 16 output layers, one layer for each depth. Formally, let a neural network $\mathcal{N}$ be parameterized by depth $\mathcal{D}$ as in Eq.~\eqref{eq:depth}.

\begin{equation}
\mathcal{D} \in \{1, 2, 3, \dots, \mathcal{D}_{max}\}
\label{eq:depth}
\end{equation}

where \(\mathcal{D}_{max}\) denotes the maximum depth (total number of blocks) in the network.

\subsection{Warm-Up with Full Depth}
\begin{figure}[t]
    \centering
    \begin{subfigure}[t]{0.41\textwidth}
        \centering
        \includegraphics[width=\textwidth]{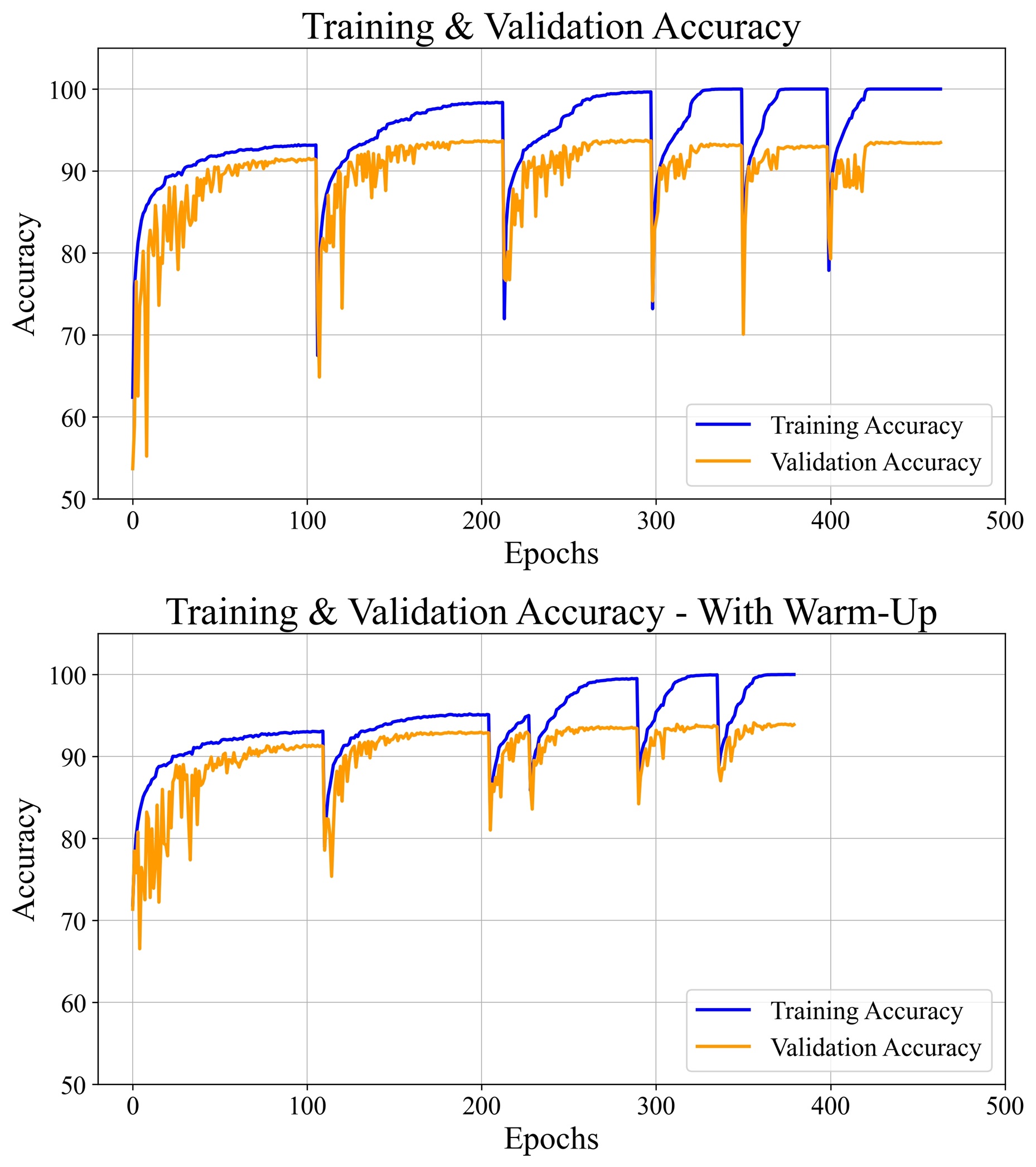}
        \caption{\centering \phantom{*}}
        \label{fig:curve_a}
    \end{subfigure}
    \hfill
    \begin{subfigure}[t]{0.41\textwidth}
        \centering
        \includegraphics[width=\textwidth]{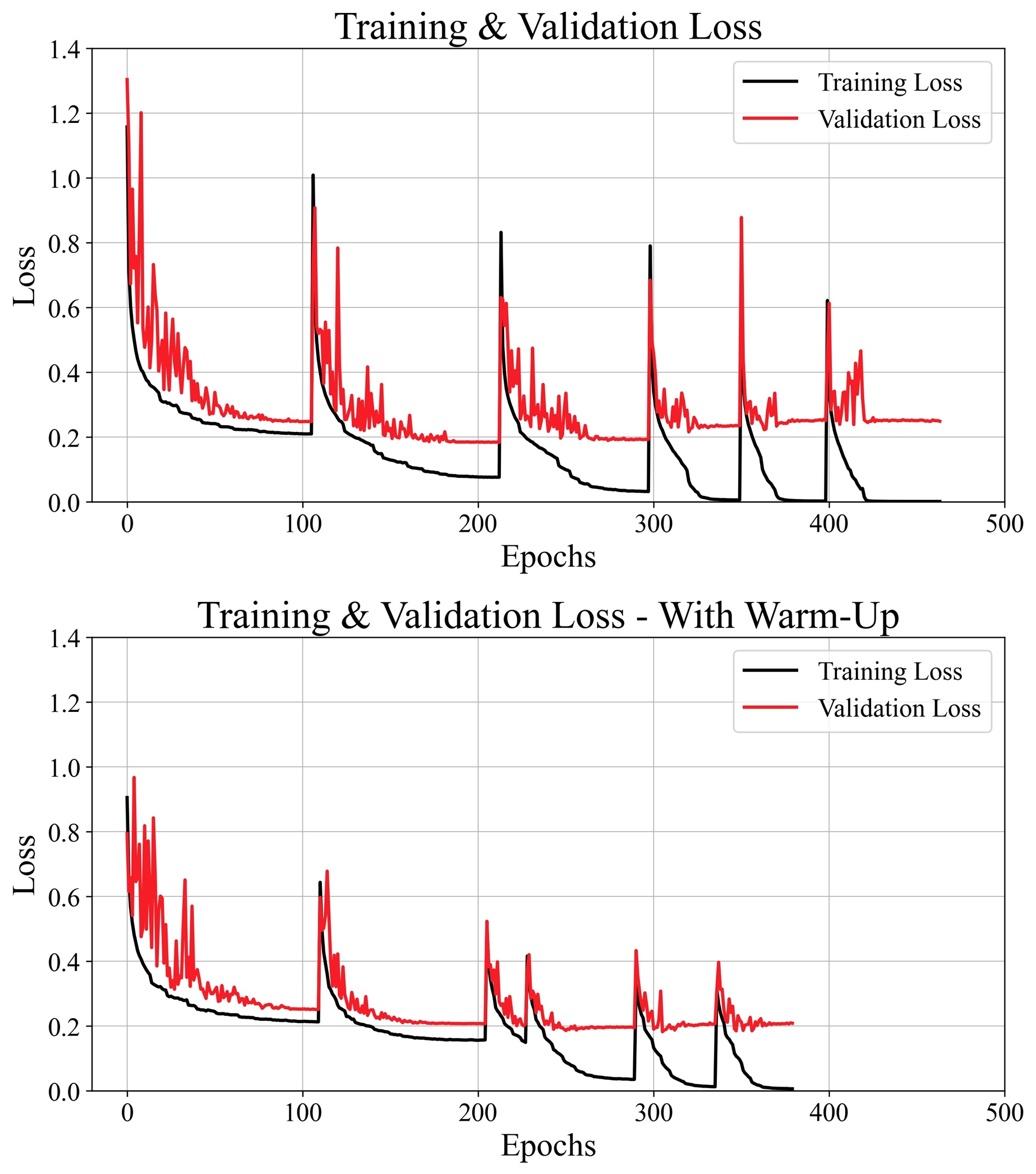}
        \caption{\centering \phantom{*}}
        \label{fig:curve_b}
    \end{subfigure}
    \caption{\centering The accuracy and loss curves of ResNet-34 for Fashion-MNIST show that use of warm-up mitigates large fluctuations and accelerates convergence during training.}
    \label{fig:figure_4}
    \vspace{-0.20cm}
\end{figure}

After depth partitioning and assignment of separate output layers for each depth, the model is warmed up at all the depths for a few epochs with a small learning rate, see Figure~\ref{fig:figure_3}(a). This gives all depth levels good initialization points, stabilizing the progressive depth expansion based training, preventing the parameters of newly activated deeper blocks from random initialization that leads to jumps in the loss curve, gradient instability, and slower convergence as shown in Figure~\ref{fig:figure_4}.

\subsection{Training with Progressive Depth Expansion}
Progressive depth expansion based training starts from the shallow portion of the network that comprises initial block(s) up to a depth level. Once the shallow block(s) have converged, next block is appended to increase the model depth and training is resumed. At each depth increment, training is started from the warmed-up state of the model instead of starting from scratch. This progressive depth expansion continues until either the target accuracy is achieved or the model reaches its maximum depth. If the target accuracy is achieved using a depth smaller than the full model depth, the current depth is declared as the optimal depth, and the ODN is extracted with the corresponding output layer for fine-tuning as shown in Figure~\ref{fig:figure_3}(b). The training objective with model depth \enquote{\(\mathcal{D}\)} is to minimize a standard supervised learning loss function:

\begin{equation}
\mathcal{L}(\Theta_{\mathcal{D}}) = \frac{1}{N} \sum_{i=1}^{N} \mathcal{L}\big(\mathcal{N}(x_{i}; \Theta_{\mathcal{D}}), \, y_{i}\big)
\label{eq:loss}
\end{equation}

where $\mathcal{L}$ is cross entropy loss, \(\Theta_{\mathcal{D}}\) represents the parameters of the network $\mathcal{N}$ activated till depth \(\mathcal{D}\), $N$ is the number of training samples in a mini-batch, $x_{i}$ represents input data and $y_{i}$ denotes corresponding ground-truth labels.  
Traditionally, a model is always trained using depth \(\mathcal{D} = \mathcal{D}_{max}\), regardless of the dataset’s complexity.  
The goal of ODNs is to optimally adapt the depth \(\mathcal{D}_{opt} \leq \mathcal{D}_{max}\) of the model for the task at hand, bridging the gap between performance and efficiency.

\vspace{0.20cm}
\textbf{Stopping Criterion:} Each depth is trained until convergence and the convergence is determined using a stopping criterion. During training or fine-tuning, the learning rate is reduced by 60 \% after every 5 consecutive epochs with \enquote{no-improvement} on the validation set. Whenever a new best validation accuracy is achieved, the counter is reset to zero. The model is considered converged when the \enquote{no-improvement} epochs reach the value of 23, as shown in Algorithm~\ref{alg:algo_1}.

\vspace{0.15cm}
\subsection{Fine-tuning the Optimally Deep Network}
Once the Optimally Deep Network is obtained, it is fine-tuned till convergence to maximize its performance as shown in Figure~\ref{fig:figure_3}(c). The fine-tuning stage is the final step in training Optimally Deep Networks (ODNs) to ensure the selected optimal depth achieves its best possible performance. Fine-tuning allows the model parameters with optimal depth to adjust more precisely to the target dataset, further improving accuracy (without altering the depth). The ODN can be extracted from the backbone at the end (by copying weights) for deployment.

\vspace{0.15cm}
\section{Experiments \& Results}
We evaluate Optimally Deep Networks (ODNs) across five benchmark datasets of varying complexity: MNIST, EMNIST, SVHN, Fashion-MNIST, and CIFAR-10. All the datasets have 10 classes except EMNIST, which has 26 classes (for \enquote{letters}). To demonstrate the simplicity and applicability of our approach, we use three widely used architectures: ResNet-18, ResNet-34, and ResNet-50. We train the models using SGD optimizer with momentum 0.9, weight decay 5e-4, batch size 128, and initial learning rate of 0.1.

\vspace{0.30cm}
Table~\ref{tab1} presents the performance of ODNs as compared to networks with full depths. It can be observed that the optimally deep ResNet-18 trained on MNIST and EMNIST achieves a comparable accuracy of 99.31 \% and 95.04 \% while reducing the model size (memory footprint) by 98.64 \% and 96.54 \%, respectively. Only 2 out of the 8 blocks of ResNet-18 are sufficient to match the complexity of the MNIST dataset whereas only 3 blocks are sufficient for EMNIST. Similarly, the optimally deep ResNet-34 achieves accuracies of 96.08 \% and 93.54 \%, with model size reductions of 96.44 \% and 95.04 \% on SVHN and Fashion-MNIST, respectively. For the CIFAR-10 dataset, the optimally deep ResNet-50 with a depth of 11 blocks instead of 16, provides the accuracy of 94.12 \% with a reduction of 73.06 \% in model size. Compared to full-depth models, the accuracies of ODNs can be achieved within \textbf{1\% tolerance}, while the memory footprint reductions are substantial, \textbf{exceeding 95 \%} for low-complexity datasets: MNIST, EMNIST, SVHN, and Fashion-MNIST.

\vspace{0.20cm}
Table~\ref{tab1} also shows that the substantial reduction in parameter count and model size of ODNs leads to a decrease in FLOPs. Lower FLOPs result in faster inference times and reduced latency in edge devices. For MNIST and EMNIST, the memory footprint is reduced from 44.78 MB to 0.61 MB and 1.55 MB, respectively. Similarly, for SVHN and Fashion-MNIST, the memory footprint decreases from 85.29 MB to 3.04 MB and 4.23 MB, respectively. In the case of CIFAR-10, the memory footprint is reduced by 73.06 \%, from 94.43 MB to 25.44 MB, reflecting the higher complexity of this dataset.

\vspace{0.20cm}
ODNs eliminate the need for an expensive and complex search space as compared to complex NAS methods. Unlike sparsification and pruning, ODNs achieve a substantial reduction in the memory footprints of the models. In contrast to Once-for-All (OFA) training, ODNs are simpler, requiring only a single progressively expanding model without the complexity of maintaining a super-net \cite{OFA_1}. Another advantage of ODNs is that in order to find the optimal depth, the model is trained till convergence using shallower depths during progressive depth expansion stage. These fully converged intermediate models can be saved as by-products at each depth and later used for deployment on devices (like OFA) when stricter accuracy-efficiency trade-offs are required.

\vspace{0.10cm}
\section*{Conclusion}
We propose Optimally Deep Networks (ODNs), models which adapt their depth according to the complexity of the task at hand. Typically, powerful architectures are trained at full depths but we show that not all datasets require high model capacity. ODNs are trained using progressive depth expansion: training begins with shallow depths and the network is incrementally deepened only as needed, until the desired performance is achieved. This enables the extraction of Optimally Deep Networks that significantly reduce memory footprint, computational overhead, and inference cost, while maintaining competitive accuracy. Empirical evaluations on MNIST, EMNIST, SVHN, Fashion-MNIST, and CIFAR-10 using ResNet-18, ResNet-34, and ResNet-50 show that ODNs can achieve up to 98.64 \% reduction in model size while maintaining good performance. ODNs present a practical framework for searching optimally deep networks that balance resource efficiency with accuracy, paving the way for scalable and deployable neural networks in real-world applications.

\vspace{0.10cm}
\bibliographystyle{IEEEtran}
\bibliography{references} 

@String(ECCV= {Eur. Conf. Comput. Vis.})

@String(ICASSP=	{ICASSP})

@String(AAAI = {AAAI})

@String(ECCV  = {ECCV})

@article{KD_1,
  title={Distilling the knowledge in a neural network},
  author={Hinton, Geoffrey and Vinyals, Oriol and Dean, Jeff},
  journal={arXiv preprint arXiv:1503.02531},
  year={2015}
}

@inproceedings{KD_2,
  title={Adversarially robust distillation},
  author={Goldblum, Micah and Fowl, Liam and Feizi, Soheil and Goldstein, Tom},
  booktitle={Proceedings of the AAAI conference on artificial intelligence},
  volume={34},
  number={04},
  pages={3996--4003},
  year={2020}
}

@article{DL_1,
  title={Deep learning},
  author={LeCun, Yann and Bengio, Yoshua and Hinton, Geoffrey},
  journal={nature},
  volume={521},
  number={7553},
  pages={436--444},
  year={2015},
  publisher={Nature Publishing Group UK London}
}

@article{KD_3,
  title={Knowledge distillation: A survey},
  author={Gou, Jianping and Yu, Baosheng and Maybank, Stephen J and Tao, Dacheng},
  journal={International Journal of Computer Vision},
  volume={129},
  number={6},
  pages={1789--1819},
  year={2021},
  publisher={Springer}
}

@misc{KD_4,
	title = {Distillation as a {Defense} to {Adversarial} {Perturbations} against {Deep} {Neural} {Networks}},
	url = {http://arxiv.org/abs/1511.04508},
	urldate = {2024-11-04},
	publisher = {arXiv},
	author = {Papernot, Nicolas and McDaniel, Patrick and Wu, Xi and Jha, Somesh and Swami, Ananthram},
	month = mar,
	year = {2016},
	note = {arXiv:1511.04508 [cs, stat]},
}

@misc{OFA_1,
	title = {Once-for-{All}: {Train} {One} {Network} and {Specialize} it for {Efficient} {Deployment}},
	shorttitle = {Once-for-{All}},
	url = {http://arxiv.org/abs/1908.09791},
	urldate = {2024-11-04},
	publisher = {arXiv},
	author = {Cai, Han and Gan, Chuang and Wang, Tianzhe and Zhang, Zhekai and Han, Song},
	month = apr,
	year = {2020},
	note = {arXiv:1908.09791 [cs, stat]},
}

@misc{OFA_2,
	title = {Once-for-{All} {Adversarial} {Training}: {In}-{Situ} {Tradeoff} between {Robustness} and {Accuracy} for {Free}},
	shorttitle = {Once-for-{All} {Adversarial} {Training}},
	url = {http://arxiv.org/abs/2010.11828},
	urldate = {2024-11-04},
	publisher = {arXiv},
	author = {Wang, Haotao and Chen, Tianlong and Gui, Shupeng and Hu, Ting-Kuei and Liu, Ji and Wang, Zhangyang},
	month = nov,
	year = {2020},
	note = {arXiv:2010.11828 [cs]},
}

@article{SNN_1,
  title={Slimmable neural networks},
  author={Yu, Jiahui and Yang, Linjie and Xu, Ning and Yang, Jianchao and Huang, Thomas},
  journal={arXiv preprint arXiv:1812.08928},
  year={2018}
}

@inproceedings{SNN_2,
  title={Universally slimmable networks and improved training techniques},
  author={Yu, Jiahui and Huang, Thomas S},
  booktitle={Proceedings of the IEEE/CVF international conference on computer vision},
  pages={1803--1811},
  year={2019}
}

@article{EDL_1,
  title={Multi-scale dense networks for resource efficient image classification},
  author={Huang, Gao and Chen, Danlu and Li, Tianhong and Wu, Felix and Van Der Maaten, Laurens and Weinberger, Kilian Q},
  journal={arXiv preprint arXiv:1703.09844},
  year={2017}
}

@inproceedings{DI_1,
  title={Skipnet: Learning dynamic routing in convolutional networks},
  author={Wang, Xin and Yu, Fisher and Dou, Zi-Yi and Darrell, Trevor and Gonzalez, Joseph E},
  booktitle={Proceedings of the European conference on computer vision (ECCV)},
  pages={409--424},
  year={2018}
}

@inproceedings{DL_2,
  title={Deep residual learning for image recognition},
  author={He, Kaiming and Zhang, Xiangyu and Ren, Shaoqing and Sun, Jian},
  booktitle={Proceedings of the IEEE conference on computer vision and pattern recognition},
  pages={770--778},
  year={2016}
}

@article{WRN,
  title={Wide residual networks},
  author={Zagoruyko, Sergey and Komodakis, Nikos},
  journal={arXiv preprint arXiv:1605.07146},
  year={2016}
}

@article{MN_1,
  title={Mobilenets: Efficient convolutional neural networks for mobile vision applications},
  author={Howard, Andrew G and Zhu, Menglong and Chen, Bo and Kalenichenko, Dmitry and Wang, Weijun and Weyand, Tobias and Andreetto, Marco and Adam, Hartwig},
  journal={arXiv preprint arXiv:1704.04861},
  year={2017}
}

@inproceedings{MN_2,
  title={Mobilenetv2: Inverted residuals and linear bottlenecks},
  author={Sandler, Mark and Howard, Andrew and Zhu, Menglong and Zhmoginov, Andrey and Chen, Liang-Chieh},
  booktitle={Proceedings of the IEEE conference on computer vision and pattern recognition},
  pages={4510--4520},
  year={2018}
}

@article{OFA_3,
  title={Only train once: A one-shot neural network training and pruning framework},
  author={Chen, Tianyi and Ji, Bo and Ding, Tianyu and Fang, Biyi and Wang, Guanyi and Zhu, Zhihui and Liang, Luming and Shi, Yixin and Yi, Sheng and Tu, Xiao},
  journal={Advances in Neural Information Processing Systems},
  volume={34},
  pages={19637--19651},
  year={2021}
}

@article{EDL_2,
  title={Progressive channel-shrinking network},
  author={Pan, Jianhong and Yang, Siyuan and Foo, Lin Geng and Ke, Qiuhong and Rahmani, Hossein and Fan, Zhipeng and Liu, Jun},
  journal={IEEE Transactions on Multimedia},
  volume={26},
  pages={2016--2026},
  year={2023},
  publisher={IEEE}
}

@inproceedings{KD_5,
  title={Confidence-aware multi-teacher knowledge distillation},
  author={Zhang, Hailin and Chen, Defang and Wang, Can},
  booktitle={ICASSP 2022-2022 IEEE International Conference on Acoustics, Speech and Signal Processing (ICASSP)},
  pages={4498--4502},
  year={2022},
  organization={IEEE}
}

@incollection{DL_3,
  title={Evolving deep neural networks},
  author={Miikkulainen, Risto and Liang, Jason and Meyerson, Elliot and Rawal, Aditya and Fink, Dan and Francon, Olivier and Raju, Bala and Shahrzad, Hormoz and Navruzyan, Arshak and Duffy, Nigel and others},
  booktitle={Artificial intelligence in the age of neural networks and brain computing},
  pages={269--287},
  year={2024},
  publisher={Elsevier}
}

@article{MI_1,
  title={Deep learning for biological image classification},
  author={Affonso, Carlos and Rossi, Andr{\'e} Luis Debiaso and Vieira, F{\'a}bio Henrique Antunes and de Leon Ferreira, Andr{\'e} Carlos Ponce and others},
  journal={Expert systems with applications},
  volume={85},
  pages={114--122},
  year={2017},
  publisher={Elsevier}
}

@article{II_1,
  title={Automatic fruit classification using deep learning for industrial applications},
  author={Hossain, M Shamim and Al-Hammadi, Muneer and Muhammad, Ghulam},
  journal={IEEE transactions on industrial informatics},
  volume={15},
  number={2},
  pages={1027--1034},
  year={2018},
  publisher={IEEE}
}

@article{DC_1,
  title={CNN model for image classification on MNIST and fashion-MNIST dataset},
  author={Kadam, Shivam S and Adamuthe, Amol C and Patil, Ashwini B},
  journal={Journal of scientific research},
  volume={64},
  number={2},
  pages={374--384},
  year={2020},
  publisher={Banaras Hindu University}
}

@article{CC_1,
  title={Deep learning--based text classification: a comprehensive review},
  author={Minaee, Shervin and Kalchbrenner, Nal and Cambria, Erik and Nikzad, Narjes and Chenaghlu, Meysam and Gao, Jianfeng},
  journal={ACM computing surveys (CSUR)},
  volume={54},
  number={3},
  pages={1--40},
  year={2021},
  publisher={ACM New York, NY, USA}
}

@inproceedings{SN_1,
  title={Siamese neural networks for one-shot image recognition},
  author={Koch, Gregory and Zemel, Richard and Salakhutdinov, Ruslan and others},
  booktitle={ICML deep learning workshop},
  volume={2},
  number={1},
  pages={1--30},
  year={2015},
  organization={Lille}
}

@article{pruning,
  title={Learning both weights and connections for efficient neural network},
  author={Han, Song and Pool, Jeff and Tran, John and Dally, William},
  journal={Advances in neural information processing systems},
  volume={28},
  year={2015}
}

@article{quantization,
  title={Quantized neural networks: Training neural networks with low precision weights and activations},
  author={Hubara, Itay and Courbariaux, Matthieu and Soudry, Daniel and El-Yaniv, Ran and Bengio, Yoshua},
  journal={journal of machine learning research},
  volume={18},
  number={187},
  pages={1--30},
  year={2018}
}

@article{NAS_1,
  title={Neural architecture search: A survey},
  author={Elsken, Thomas and Metzen, Jan Hendrik and Hutter, Frank},
  journal={Journal of Machine Learning Research},
  volume={20},
  number={55},
  pages={1--21},
  year={2019}
}

@inproceedings{NAS_2,
  title={Progressive neural architecture search},
  author={Liu, Chenxi and Zoph, Barret and Neumann, Maxim and Shlens, Jonathon and Hua, Wei and Li, Li-Jia and Fei-Fei, Li and Yuille, Alan and Huang, Jonathan and Murphy, Kevin},
  booktitle={Proceedings of the European conference on computer vision (ECCV)},
  pages={19--34},
  year={2018}
}

@inproceedings{NAS_3,
  title={Efficient neural architecture search via parameters sharing},
  author={Pham, Hieu and Guan, Melody and Zoph, Barret and Le, Quoc and Dean, Jeff},
  booktitle={International conference on machine learning},
  pages={4095--4104},
  year={2018},
  organization={PMLR}
}

@inproceedings{RENAS,
  title={Renas: Reinforced evolutionary neural architecture search},
  author={Chen, Yukang and Meng, Gaofeng and Zhang, Qian and Xiang, Shiming and Huang, Chang and Mu, Lisen and Wang, Xinggang},
  booktitle={Proceedings of the IEEE/CVF conference on computer vision and pattern recognition},
  pages={4787--4796},
  year={2019}
}

@article{NPENAS,
  title={Npenas: Neural predictor guided evolution for neural architecture search},
  author={Wei, Chen and Niu, Chuang and Tang, Yiping and Wang, Yue and Hu, Haihong and Liang, Jimin},
  journal={IEEE transactions on neural networks and learning systems},
  volume={34},
  number={11},
  pages={8441--8455},
  year={2022},
  publisher={IEEE}
}

@article{DARTS,
  title={Darts: Differentiable architecture search},
  author={Liu, Hanxiao and Simonyan, Karen and Yang, Yiming},
  journal={arXiv preprint arXiv:1806.09055},
  year={2018}
}
\end{document}